\documentclass{article}
\usepackage{graphicx} 
\usepackage[skip=10pt plus1pt, indent=5pt]{parskip}

\usepackage{amsfonts}
\usepackage{amsmath}
\usepackage{amssymb}
\usepackage{ntheorem}
\usepackage{authblk}
\usepackage[square,sort,comma,numbers]{natbib}
\usepackage[hidelinks]{hyperref}
\usepackage{thm-restate}

\theoremstyle{break}

\newtheorem{proof}{Proof}

\theorembodyfont{}
\newtheorem{definition}{Definition}

\newcommand{\I}{\mathcal{I}}
\newcommand{\C}{\mathcal{C}}

\newcommand{\E}{\mathbb{E}}

\newcommand{\pairs}{\mathcal{P}}
\newcommand{\pairNum}{N}

\newcommand{\pointNum}{|\mathcal{I}|}
\newcommand{\clusterNum}{n}

\newcommand{\perm}{Perm }
\newcommand{\num}{Num }
\newcommand{\all}{All }
\newcommand{\fit}{Fit }
\newcommand{\sym}{Sym }
\newcommand{\flatt}{Flat }

\newcommand{\RI}
{\mathcal{R}\mathcal{I}}

\newcommand{\ARI}
{\mathcal{A}\mathcal{R}\mathcal{I}}
\newcommand{\conc}{\mathbf{conc}}
\newcommand{\simp}{\Delta}
\newcommand{\UHC}[1]{[0, 1]^{#1}}
\newcommand{\textdef}[1]{\textbf{#1}}
\newcommand{\id}[1]{\Tilde{#1}}

\newcommand\blfootnote[1]{%
  \begingroup
 \renewcommand\thefootnote{}\footnote{#1}%
  \addtocounter{footnote}{-1}%
  \endgroup
}

\title{Random Models for Fuzzy Clustering Similarity Measures}
\date{}

\author[1]{Ryan DeWolfe}
\author[1]{Jeffrey L. Andrews}
\affil[1]{University of British Columbia – Okanagan Campus,
Kelowna, BC, Canada}

\begin{document}

\maketitle

\blfootnote{Support for this work was provided by the Natural Sciences and Engineering Research Council of 
Canada Discovery Grants program (RGPIN-2020-04646, Dr. Jeffrey L. Andrews)}

\begin{abstract}
    The Adjusted Rand Index (ARI) is a widely used method for comparing hard clusterings, but requires a choice of random model that is often left implicit.
    Several recent works have extended the Rand Index to fuzzy clusterings, but the assumptions of the most common random model is difficult to justify in fuzzy settings. 
    We propose a single framework for computing the ARI with three random models that are intuitive and explainable for both hard and fuzzy clusterings, along with the benefit of lower computational complexity.
    The theory and assumptions of the proposed models are contrasted with the existing permutation model.
    Computations on synthetic and benchmark data show that each model has distinct behaviour, meaning that accurate model selection is important for the reliability of results.
\end{abstract}

\section{Introduction}

In unsupervised clustering, where true labels do not technically exist and accuracy is poorly defined, it is still useful to compare the results of several cluster analyses. 
These techniques are called external comparison/validity indices and are used to measure the similarity of two clustering results.
Clustering algorithm performance is often reported in terms of the ability to produce a clustering that is similar to some known clustering that is treated as correct.
With the variety of external comparison indices, each having particular assumptions and behaviours, it is not rare to see several used in simulation studies without any attempt to indicate which index is the most meaningful for the situation.

A popular external comparison index is the Rand index \citep{rand1971}, the proportion of pairs of points that both clustering algorithms agree on the answer to: are these points in the same cluster?
The maximum of the Rand index is 1, and is obtained by identical clustering results (up to a permutation of labels).
However, even cluster allocations performed at random often have large Rand index values, making it difficult to determine if clustering algorithms are finding similar structures.
To rectify this, the Rand index was adjusted for chance \citep{hubert1985ARI} via

\begin{equation}\label{eq:ai}
    \ARI = \frac{\RI -\E_{model}[\RI]}{\max[\RI] - \E_{model}[\RI]}.
\end{equation}
Here, a value of 1 still represents maximum similarity, but 0 now represents the level of expected similarity if the cluster allocations had been chosen at random.
Of course `at random' is not well defined, and several models for adjustment have been suggested in the hard clustering literature (\citep{gates2017models}, \citep{hubert1985ARI}, \citep{morey1984ARI}).

Some clustering approaches, such as mixture model-based (\citep{mclachlan2019finite}, \citep{mcnicholas2016mixture}), produce fuzzy or probabilistic cluster allocations.
In these cases, points can partially belong to several clusters, allowing for the algorithm to provide uncertainty in the allocations.
This type of clustering result also benefits from external comparison indices, and there have been several extensions of the Rand Index to fuzzy cluster allocations (\citep{andrews2022assessments}, \citep{brouwer2009}, \citep{campagner2023frameworkFuzzyIndices}, \citep{hullermeier2011comparing}).
One random model --- the permutation model --- has been used as a fuzzy extension to the adjusted Rand index (\citep{andrews2022assessments}, \citep{d2021adjusted}), but in the fuzzy realm where cluster sizes are not well defined, the assumptions of this model become muddled and difficult to justify.

In this paper, we propose a unified approach to extending the original multinomial model \citep{morey1984ARI} to fuzzy allocations in such a way the it can mimic two of the hard adjustment models \citep{gates2017models}, along with assumptions that transfer cleanly to fuzzy clustering.
Section \ref{sec:hardModels} defines the Rand index and examines previously proposed models for hard clustering.
Some extensions of the Rand index to probabilistic clustering are summarized in Section \ref{sec:FuzzyRandExtensions}.
Our models, along with some theory, are presented in Section \ref{sec:dirModels} with benchmarks and simulations in Section \ref{sec:benchAndSim}.
Finally, we summarize the performance of our models and suggest some directions for future work in Section 6.

\section{Hard Random Models}\label{sec:hardModels}

\subsection{The Rand Index}

\begin{definition}[Clustering]
    Let $\I$ be an index set of the points  being clustered.
    A clustering of $\I$ into up to $\clusterNum$ clusters is a function $\C: \I \to \UHC{n}$.

    \begin{itemize}
        \item Any clustering can be considered \textdef{possibilistic}.
        \item $\C$ is a \textdef{fuzzy or probabilistic} clustering if $Image(\C) \subset \simp^{\clusterNum-1}$. The $\clusterNum-1$ dimensional \textdef{canonical simplex} $\simp^{\clusterNum-1}$ is the set of vectors in $\UHC{n}$ whose coordinates sum to 1.
        \item $\C$ is a \textdef{hard} clustering if $\forall i \in \I, \C(i)$ contains one $1$ and the rest are 0.
        \end{itemize}
\end{definition}

The Rand index \citep{rand1971} is a measure of similarity between two hard clustering results with a maximum of $1$ denoting perfect similarity.
There are many equivalent formulations of the Rand index so we will provide the original equation and a reformulation that will be useful in Sections \ref{sec:FuzzyRandExtensions} and \ref{sec:dirModels}.

Let $\pairs = \{ \{p_1, p_2\} | p_1 < p_2 \: \forall \: p_1, p_2 \in \I\}$ be the unordered two element subsets of $\I$.
Let $\C_1$ and $\C_2$ be hard cluster allocations.
Partition $\pairs$ as follows:

\begin{itemize}
    \item $p \in \mathbf{A}$ if  $\C_1(p_1) = \C_1(p_2)$ and $\C_2(p_1) = \C_2(p_2)$.
    \item $p \in \mathbf{B}$ if $\C_1(p_1) = \C_1(p_2)$ and $\C_2(p_1) \neq \C_2(p_2)$.
    \item $p \in \mathbf{C}$ if $\C_1(p_1) \neq \C_1(p_2)$ and $\C_2(p_1) = \C_2(p_2)$.
    \item $p \in \mathbf{D}$ if $\C_1(p_1) \neq \C_1(p_2)$ and $\C_2(p_1) \neq \C_2(p_2)$.
\end{itemize}

\begin{definition}[Rand Index]
    \begin{equation}
        \RI(\C_1, \C_2) = \frac{ |\mathbf{A}| + |\mathbf{D}| }{ |\mathbf{A}| + |\mathbf{B}| + |\mathbf{C}| + |\mathbf{D}| } = \frac{ |\mathbf{A}| + |\mathbf{D}| }{ {|\I| \choose 2} }
    \end{equation}
\end{definition}

The set $A \cup D$ is called the \textdef{concordant pairs} and $B \cup C$ the \textdef{discordant pairs}.
A pair $p$ is concordant if $\C_1$ and $\C_2$ have the same answer to the question: \textit{Are $p_1$ and $p_2$ in the same cluster?} This is the context of the original formulation \citep{rand1971}.

In an equivalent formulation we require two functions to sort the concordant pairs.
The first in an intra-clustering agreement function that will measure whether two points are placed into the same cluster.
The second is an inter-cluster concordance function which will measure the similarity of the agreement value from the two clustering results. 

\begin{definition}[Agreement]
    An agreement function $a_{C}:\pairs \to [0, 1]$ is any function such that
    \begin{enumerate}
        \item $\C(p_1) = C(p_2) \implies a_\C(p) = 1$
        \item $\C(p_1) \neq C(p_2) \implies a_\C(p) = 0$
    \end{enumerate}
\end{definition}

\begin{definition}[Concordance]\label{def:hardConc}
    \begin{equation}
        \conc_{\C_1, \C_2}(p) = 
        \begin{cases}
            1 & a_{\C_1}(p) = a_{\C_2}(p)\\
            0 & a_{\C_1}(p) \neq a_{\C_2}(p)\\
        \end{cases}
    \end{equation}
\end{definition}

In general we will exclude the subscripts where the clustering(s) under consideration are arbitrary or evident given the situation.

\begin{definition}[Equivalent Rand Index]\label{def:ri}
An equivalent formulation of the Rand index is the expectation of the concordance function over a uniform distribution on the set $\pairs$.
    \begin{equation}
        \RI = \E[\conc]
    \end{equation}
\end{definition}

While both formulations are identical for hard cluster allocations, the agreement and concordance functions can be easily extended to consider fuzzy membership vectors and less than perfect agreement.
Several such extensions have been proposed and will be reviewed in Section \ref{sec:FuzzyRandExtensions}.

\subsection{Permutation Model}
The most common model for adjusting the Rand index is the permutation model, proposed in by Hubert and Arabie in 1985 \citep{hubert1985ARI}.
A random clustering is selected by applying a random permutation to the set $\I$ selected with uniform probability.
This random model has the effect of fixing both the number and size of clusters to those in $\C_1$ and $\C_2$.

Given hard clustering $\C_1$, let $\mathbf{c}^{(1)}$ be a vector of length $\clusterNum$, with each $c_i$ counting the number of points with the one in dimension $i$ from $\C_1$.
Similarly define $\mathbf{c}^{(2)}$ given $\C_2$.
Let $\pairNum = |\pairs| = {|\I| \choose 2}$ be the number of pairs.
The expectation of the rand index under the permutation model is as follows.

\begin{equation}\label{eq:eriperm}
    \E_{Perm}[\RI] = \frac{\sum_i {c^{(1)}_i \choose 2} * \sum_i {c^{(2)}_i \choose 2} + \left (\pairNum - \sum_i {c^{(1)}_i \choose 2} \right) *\left(\pairNum - \sum_i {c^{(2)}_i \choose 2}\right)}{\pairNum}
\end{equation}

This model is widely used and many properties have been explored in previous papers (\citep{steinley2004properties}, \citep{steinley2018noteonrand}, \citep{steinley2016variance}, \citep{warrens2022understanding}).
However, the model is quite restrictive and the assumptions are not always reasonable.
For example, the K-means algorithm \citep{jain2010data} fixes the number, but not the size, of clusters.
Thus, the permutation model does not consider all possible cluster allocations that could have reasonably been produced by the K-means algorithm, as mentioned in \citep{gates2017models}.

\subsection{Categorical/Multinomial}
The first proposed random model was presented in 1984 \citep{morey1984ARI} and assumed the cluster size vector $\mathbf{c}$ follows a multinomial distribution.
This model is usually presented in terms of the multinomial random variable, but for following sections it is useful to decompose it into a sum of categorical random variables.
For consistency, first let us define a categorical random variable.

\begin{definition}[Categorical Random Variable]
    A categorical random variable $Cat$ with $\clusterNum$ categories, parameterized by a vector $p_i \in \simp^{\clusterNum - 1}$ has the following distribution.

    $P(Cat = e_i) = p_i$ where $e_i$ is a $\clusterNum$ dimensional vector with a $1$ in the i\textsuperscript{th} dimension and $0$ elsewhere.
\end{definition}

Given cluster allocations $\C_1$ and $\C_2$ with $n_1$ and $n_2$ number of clusters respectively, categorical variables $Cat_1$ and $Cat_2$ are parameterized by cluster proportion vectors $p_1 = \frac{\mathbf{c}^{(1)}}{|\I|}$ and $p_2 = \frac{\mathbf{c}^{(2)}}{|\I|}$.
Pairs of random cluster allocations are generated by assigning each $i \in \I$ an independently sampled membership vector from $Cat_1$ and $Cat_2$.
The expectation can then be calculated.

\begin{equation} \label{eq:ericat}
    \E_{Cat}[\RI] = \sum_{i=1}^{\clusterNum_1} \sum_{j = 1}^{\clusterNum_2} \left( p^{(1)}_i \right) * \left( p^{(2)}_j \right) + \left( 1 - p^{(1)}_i \right) * \left( 1 - p^{(2)}_j \right)
\end{equation}

Every pair of cluster allocations with up to $n_1$ and $n_2$ clusters is present in this model, but the probability is not evenly distributed.
Up to a permutation of cluster labels, it is more likely that cluster allocations with close to $n_1$ and $n_2$ clusters will be chosen.
The assumptions are less rigid, but the proportion vectors $p$ ensures that randomly chosen allocations are likely to have similar number and size of clusters to the originals.

Written in this format, it is clear that what Morey and Agresti were considering is that the points are randomly (independently) assigned clusters.
Each point's random membership vector being independent is a powerful feature that is unique to this model.
In Section \ref{sec:dirModels}, we will leverage this independence to simplify the expectation calculation.
While this model is significantly less used than the permutation model, it does have some studies comparing its performance to the permutation model (\citep{steinley2018noteonrand}, \citep{sundqvist2023mri}).

\subsection{Num and All}
To relax the assumptions of the permutation model, Gates and Ahn \citep{gates2017models} proposed two more random allocation models.
In the first model, \textdef{\num}, a random clustering is selected from $\C$ by uniformly selecting from all possible cluster allocations (up to a permutation of labels) of $\I$ into $\clusterNum$ clusters.
In this model, the number of clusters is assumed to match the number of clusters observed, but the cluster sizes can vary.

\begin{equation}\label{eq:eriNum}
    \E_{\num}[\RI] = \frac{S(\pairNum-1, \clusterNum_1)}{S(\pairNum, \clusterNum_1)} \frac{S(\pairNum-1, \clusterNum_2)}{S(\pairNum, \clusterNum_2)} + \left(1 - \frac{S(\pairNum-1, \clusterNum_1)}{S(\pairNum, \clusterNum_1)} \right ) \left ( 1-\frac{S(\pairNum-1, \clusterNum_2)}{S(\pairNum, \clusterNum_2)} \right )
\end{equation}

\begin{equation}\label{eq:eriNumApprox}
    \E_{\num}[\RI] \approx \frac{1}{\clusterNum_1 * \clusterNum_2} + \left( 1 - \frac{1}{\clusterNum_1} \right) \left( 1 - \frac{1}{\clusterNum_2} \right)
\end{equation}

Where $S(\pointNum, \clusterNum)$ is the Stirling number of the second kind, the number of ways to partition a set of size $\pointNum$ into $\clusterNum$ clusters.
Stirling numbers become difficult to compute with large $\clusterNum$, so Gates and Ahn suggest approximating $S$ which is used is Equation \ref{eq:eriNumApprox}.

The second model is \textdef{All}, where a clustering is selected uniformly from every possible clustering of $\I$ (implicitly into up to $|\I|$ clusters):
\begin{equation}\label{eq:eriAll}
\E_{All}[\RI] = \left( \frac{B_{\pairNum-1}}{B_\pairNum} \right) ^2 + \left( 1 - \frac{B_{\pairNum-1}}{B_\pairNum} \right)^2
\end{equation}
Where $B_\pairNum$ is the Bell number $B_\pairNum = \sum_{i=1}^\pairNum S(\pairNum, i)$. 

These models, along with the permutation model, provide varying levels of assumptions that consider different sets of random clusters.
Shown above are the two-sided adjustments where both clustering results are randomized.
A common use for similarity indices is benchmarking, where one of the allocations is held known and fixed. In these cases, one-sided models are considered advantageous.
Gates and Ahn provide one sided formulas for \num and \all, but to our knowledge a one sided categorical model has not been previously studied.

\section{Fuzzy Rand Extensions}\label{sec:FuzzyRandExtensions}

Several clustering algorithms, such as mixture models, result in fuzzy cluster allocations to reflect uncertainty or overlap.
As these results also benefit from comparison measures, there are several extensions of the Rand index to handle fuzzy clustering results.
Since hard allocations are quite limited, and the Rand index has many equivalent formulations when extended to the probabilistic space.
We consider a fuzzy Rand extension to be an \textdef{Agreement-Concordance Type Rand Extension} if it can be written in the form of Definition \ref{def:ri} with appropriate extensions of agreement and concordance.
It is possible to create other types of fuzzy rand extensions, (\citep{andrews2022assessments},\citep{anderson2010comparing}), but our method relies on the structure of the expectation calculation.

\subsection{Brouwer}

Brouwer \citep{brouwer2009} uses the cosine similarity as the agreement function.

\begin{equation*}
a_C(p) = \frac{\C(p_1)}{||\C(p_1)||_2} \cdot \frac{\C(p_2)}{||\C(p_2)||_2}
\end{equation*}

And a concordance function
\begin{equation*}
conc(p) =  a_{\C_1}(p)a_{\C_2}(p) +  (1- a_{\C_1}(p))(1 - a_{\C_2}(p))
\end{equation*}

This choice of $a$ maintains our expected boundary behaviour and has some nice metric properties.
The $\conc$ function is an extension, but it does not have necessary properties in the interior.
If both agreements are less than 1, even if they are equal, the concordance is less than 1.
This choice of concordance makes the index not reflexive, meaning identical clustering results do not always produce the maximum index. This maximum decreases as the `fuzziness' of the allocation increases, which Andrews et al. \citep{andrews2022assessments} propose a fix for with the `Frobenius adjusted Rand index'.

\subsection{NDC and ACI}

Hullermeier et al. \citep{hullermeier2011comparing} proposed a reflexive Rand extension, the \textdef{Normalized Degree of Concordance} or NDC.
The construction works for any proper metric $||\cdot||$ on $[0, 1]^n$, but only the $\ell_1$ norm has been used in simulations.

\begin{equation}
a_\C(p) = 1 - \frac{||\C(p_1) - \C(p_2)||_1}{2}
\end{equation}

\begin{equation}
    \conc(p) = 1 - |a_{\C_1}(p) - a_{\C_2}(p)|
\end{equation}

The $\ell_1$ metric is intuitive since identical vectors always have perfect agreement.
Furthermore, if we view fuzzy membership vectors as probabilities of cluster membership, the probability that points with orthogonal membership vectors will be placed in the same cluster is $0$.

The NDC was corrected for chance using the permutation model by D'Ambrosio et al. \citep{d2021adjusted} to create the \textdef{Adjusted Concordance Index} (ACI). Andrews et al. \citep{andrews2022assessments} then provided a reasonably fast algorithm for the exact computation of the ACI with run time $O(N^2 log N)$.

\section{Dirichlet Random Models}\label{sec:dirModels}

Despite its popularity and extensive study for hard cluster allocations, the assumptions of the permutation model do not transfer well to probabilistic clustering.
For hard clustering, fixing the number and size of clusters is intuitive, but for fuzzy clustering, sizes of clusters are less well defined.
Permuting a fixed set of membership vectors is very restrictive, and many reasonable clustering results are never considered under this model.
Furthermore, it is unclear in what situation the permutation model correctly matches the assumptions in the clustering algorithm, since selecting the available set of fuzzy membership vectors before clustering would be problematic.

In this section, we will describe extending the categorical random model to the fuzzy realm using Dirichlet random variables.
The assumptions of these models are simple to explain and justify.
The underlying Dirichlet random variable will be selected to convey some assumptions about what other clustering results should be considered feasible.
First, we will develop the theory for any probability distribution on the simplex with any extended concordance functions, and then we will introduce the \textdef{Fit}, \textdef{Sym}, and \textdef{Flat} random models.

\subsection{Theory}

Suppose we are creating random cluster allocations by sampling i.i.d membership vectors from distribution $D_1, D_2$ supported on simplices $\simp_1, \simp_2$.
Let $f$ and $g$ be their respective probability density functions.
The key to these models is the random membership vectors are independent and identical samples from these distributions.
Since each pair is equally likely, we can calculate the expected concordance over the whole clustering by only considering one pair.
Furthermore, the independent membership vectors mean the 4 condition probability can be split into the product of the individual probabilities.

\begin{equation}\label{eq:ENDCIntegral}
\begin{aligned}
    &\E_{D_1, D_2}[\conc]\\
    =& \frac{1}{N} \sum_{\pairs} P\left(\C_1(p_1) = z_{11} \land \C_1(p_2) = z_{12} \land \C_2(p_1) = z_{21} \land \C_2(p_2) = z_{22}\right) \conc(z_{11}, z_{12}, z_{21}, z_{22})\\
    =& P\left(\C_1(p_1) = z_{11}\right) P\left(\C_1(p_2) = z_{12}\right) P\left( \C_2(p_1) = z_{21}\right) P\left(\C_2(p_2) = z_{22}\right) \conc(z_{11}, z_{12}, z_{21}, z_{22})\\
    =& \int_{\simp_1}\int_{\simp_1} \int_{\simp_2} \int_{\simp_2} f(z_{11})f(z_{12})g(z_{21})g(z_{22}) \conc(z_{11}, z_{12}, z_{21}, z_{22}) dz_{11}dz_{12}dz_{21}dz_{22}
\end{aligned}
\end{equation}

With equation \ref{eq:ENDCIntegral} we can compute the expected value of any Agreement-Concordance Type Rand extension given two distributions with which to create random clusterings.
Equation \ref{eq:ENDCIntegral} is generally not available in closed form, and the total dimension of integration, $2*\clusterNum_1 + 2*\clusterNum_2$, is a barrier to computation.
We will discuss the implemented and explored approximation methods in Section \ref{sec:Computation}.

\subsection{Dirichlet Models}

By replacing the random variable from the Categorical model with a Dirichlet random variable, we can easily extend the model to consider fuzzy cluster allocations. 

\begin{definition}
    A random variable $X$ with support on the the open $\clusterNum-1$ canonical simplex follows the Dirichlet distribution $X \sim Dir(\alpha)$ if the pdf is given by
    \begin{equation}
        f(x) = \frac{1}{B(\alpha)} \prod_{i=1}^n x_i^{\alpha_i - 1}
    \end{equation}
    \begin{equation}
        B(\alpha) = \frac{\prod_{i=1}^n \Gamma(\alpha_i)}{\Gamma(\sum_{i=1}^n \alpha_i)}
    \end{equation}
\end{definition}

Where $\alpha$ is a $\clusterNum$ dimensional vector with positive entries, called the concentration vector.
If all coordinates in $\alpha$ are equal, $X$ is a \textdef{Symmetric Dirichlet}, as there is no preference for any coordinate.
Finally, if all entries in $\alpha$ are 1, $X$ is a \textdef{Flat Dirichlet}.
The flat Dirichlet is the uniform probability distribution across the open simplex.

It's worth noting that this distribution will never produce a hard membership vector as part of the model.
However, since we will be integrating we do not need the function defined on the boundary.
In this sense, every possible clustering with up to $\clusterNum$ clusters will be considered, even those with fewer clusters (i.e. the probability that no points belong to the $i^{th}$ cluster more than $\varepsilon$ is non-zero).
Additionally, the limit of a sequence of Dirichlet distributions with $\alpha \to 0$ is a categorical distribution with a parameter vector $p_i = \lim \frac{\alpha_i}{\sum_i \alpha_i}$.
This makes the Dirichlet distribution a natural extension of the categorical distribution previously used for the hard case by Morey and Agresti \citep{morey1984ARI}.

We now propose the three fuzzy models.
\begin{itemize}
    \item \textdef{\fit} $D_1$ and $D_2$ are the maximum likelihood estimated Dirichlet distributions from the observed cluster allocations $C_1$ and $C_2$.
    \item \textdef{\sym} $D_1$ and $D_2$ are the maximum likelihood estimated symmetric Dirichlet distributions from the observed cluster allocations $C_1$ and $C_2$.
    \item \textdef{\flatt} $D_1$ and $D_2$ are flat Dirichlet distributions with the same number of dimensions as clusters in $C_1$ and $C_2$ respectively.
\end{itemize}

Fit is a true fuzzy extension of the hard categorical model, in the sense when the observed cluster allocations $\C_1$ and $\C_2$ are hard (or sufficiently close), the maximum likelihood estimated Dirichlet distributions will be the appropriately parameterized categorical distribution (or sufficiently close).
Compared to the previously proposed models, Fit is obviously most similar to the multinomial/categorical model, but it can also be seen as similar to the permutation model.
The assumptions in the permutation model is that random models have the same number of clusters and cluster sizes. 
While the Fit model does consider other values for the two assumptions, it is more likely that a random cluster allocation will have similar number of clusters and cluster sizes, especially as $\pointNum >> \clusterNum$. 

The Sym model makes an assumption about the number of clusters, but not about the cluster sizes.
It is more likely that the clusters will have equal sizes but that is a result of the nature of independently assigning cluster labels.
If the cluster allocations $\C_1$ and $\C_2$ are hard, the distributions will be evenly weighted categorical distributions.
Inserting $p_i = \frac{1}{\clusterNum}$ into equation \ref{eq:ericat}, we see that the expectation for \sym is the same as the approximation given for \num.
When $\pointNum >> \clusterNum$, the probability that a random clustering from \sym will have fewer than $\clusterNum$ clusters is extremely small so the distribution of models \sym and \num is quite similar, but ours can be easily extended to fuzzy cluster allocations.
The arguments proposed by Gates and Ahn \citep{gates2017models} for selecting between Num and Perm also apply to choosing between Sym and Fit, and apply seamlessly to fuzzy cluster allocations as well.

The Flat model does not have an analogous hard model, but can be considered as having the fewest amount of assumptions.
It is not well defined what the maximum number of clusters in a fuzzy scenario would be, since there could be more clusters than points, so the cluster number model parameter is necessary.
This model includes no information about preferred fuzziness nor cluster sizes.

In addition, all three models can have one-sided variants where only one of the cluster allocations is random and always compared to a fixed $C_2$.
Instead of integrating over all possible pairs of membership vectors, we sum over the observed pairs for clustering $C_2$. 

\begin{equation}\label{eq:ENDConeSideIntegral}
\begin{aligned}
    \E_{D_1, C_2}[\conc] = \frac{1}{\pairNum} \sum_{\pairs} \int_{\simp_1}\int_{\simp_1} f_1(z_{11})f_2(z_{12}) \conc \left(z_{11}, z_{12}, C_2(p_1), C_2(p_2)\right) dz_{11}dz_{12}\\
\end{aligned}
\end{equation}

\subsection{Stability}

Unfortunately, the Dirichlet distribution does not have a closed-form maximum likelihood estimator, and must be solved iteratively.
Several estimation methods are described by Minka\citep{minka2000mleDirichlet}, with up to quadratic convergence to the true parameters.
Thus, it is important that our estimations of the Dirichlet distributions used in the calculation do not have large effects on the results.
We show that the expected index is indeed stable with respect to the Dirichlet distributions used, and close estimations will provide accurate results.
The proof is presented in Appendix \ref{app:A}

\begin{restatable}[Stability]{prop}{stability}
\label{prop:endcStab}
    Let $\id{D_1}$ be an approximation of the distribution $D_1$ with support on $\simp_1$, with pdf's $\id{f}$ and $f$ respectively, such that
    \begin{equation*}
    \int_{\Delta_1} \left| \id{f}(x) - f(x) \right| dx < \frac{\varepsilon}{4}
    \end{equation*}
    Similarly for $\id{D_2}$, $D_2$, $\id{g}$, and $g$ on $\Delta_2$ such that 
    \begin{equation*}
        \int_{\simp_2} \left| \id{g}(x) - g(x)\right| dx < \frac{\varepsilon}{4}
    \end{equation*}
    Then, 
    \begin{equation}
        \left| \E_{\id{D_1}, \id{D_2}}[\conc] - \E_{D_1, D_2}[\conc] \right| < \varepsilon
    \end{equation}
\end{restatable}

\section{Benchmark and Simulation}\label{sec:benchAndSim}

\subsection{Computational Considerations}\label{sec:Computation}

The required computation for the proposed adjustment models can be considered in two parts; finding the Dirichlet parameters, and calculating the expected index. 
The first part has been studied previously in Minka \citep{minka2000mleDirichlet} and makes use of the convexity of the log-likelihood function of the Dirichlet distribution.
We implemented fixed point iterations for both Fit and Symmetric distribution estimation with a stopping condition of movement less than $10^{-10}$ or a maximum of 10 000 iterations.
Several integration techniques were tested for computing the integral from equations \ref{eq:ENDCIntegral} and \ref{eq:ENDConeSideIntegral}, including Monte Carlo, Quasi-Monte Carlo and Cubature methods.
However, since sampling from from a Dirichlet distribution is simple and extremely fast, the best performing method was to average many independent random samples from the estimated distribution.

The Dirichlet parameter estimation and index computation algorithms
\footnote{Available at \href{https://github.com/ryandewolfe33/DirichletRandAdjustmentModels}{https://github.com/ryandewolfe33/DirichletRandAdjustmentModels}.}
were implemented in Julia \citep{Julia-2017}, with distribution definitions and sampling from the Distributions.jl \citep{Distributions.jl-2019}.

\subsection{Toy Example}

\begin{figure}[!h]
    \begin{minipage}[c]{0.32\linewidth}
    \centering
    \includegraphics[width=\linewidth]{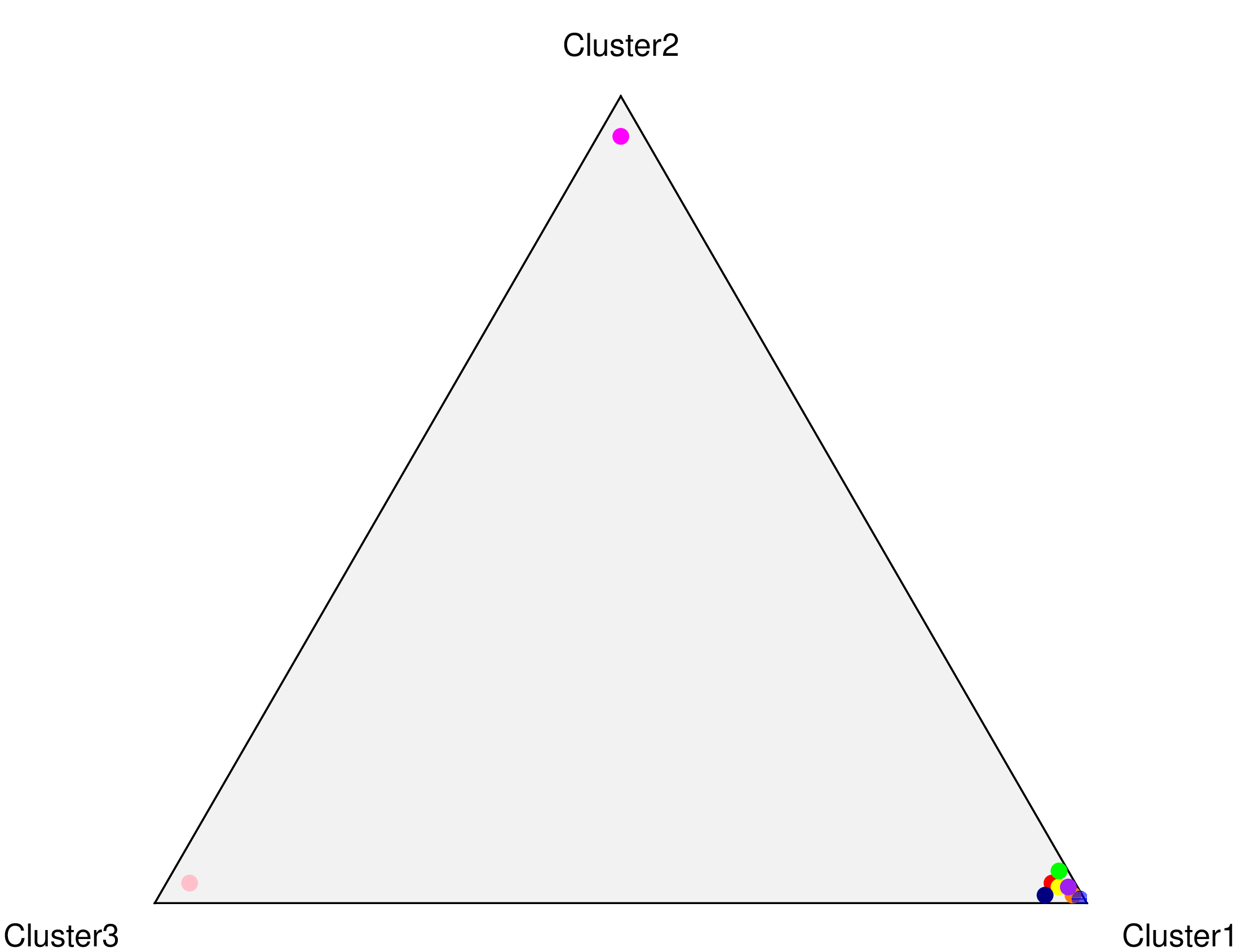}
    Uneven Low Fuzzy
    \end{minipage}
    \begin{minipage}[c]{0.32\linewidth}
        \centering
        \includegraphics[width=\linewidth]{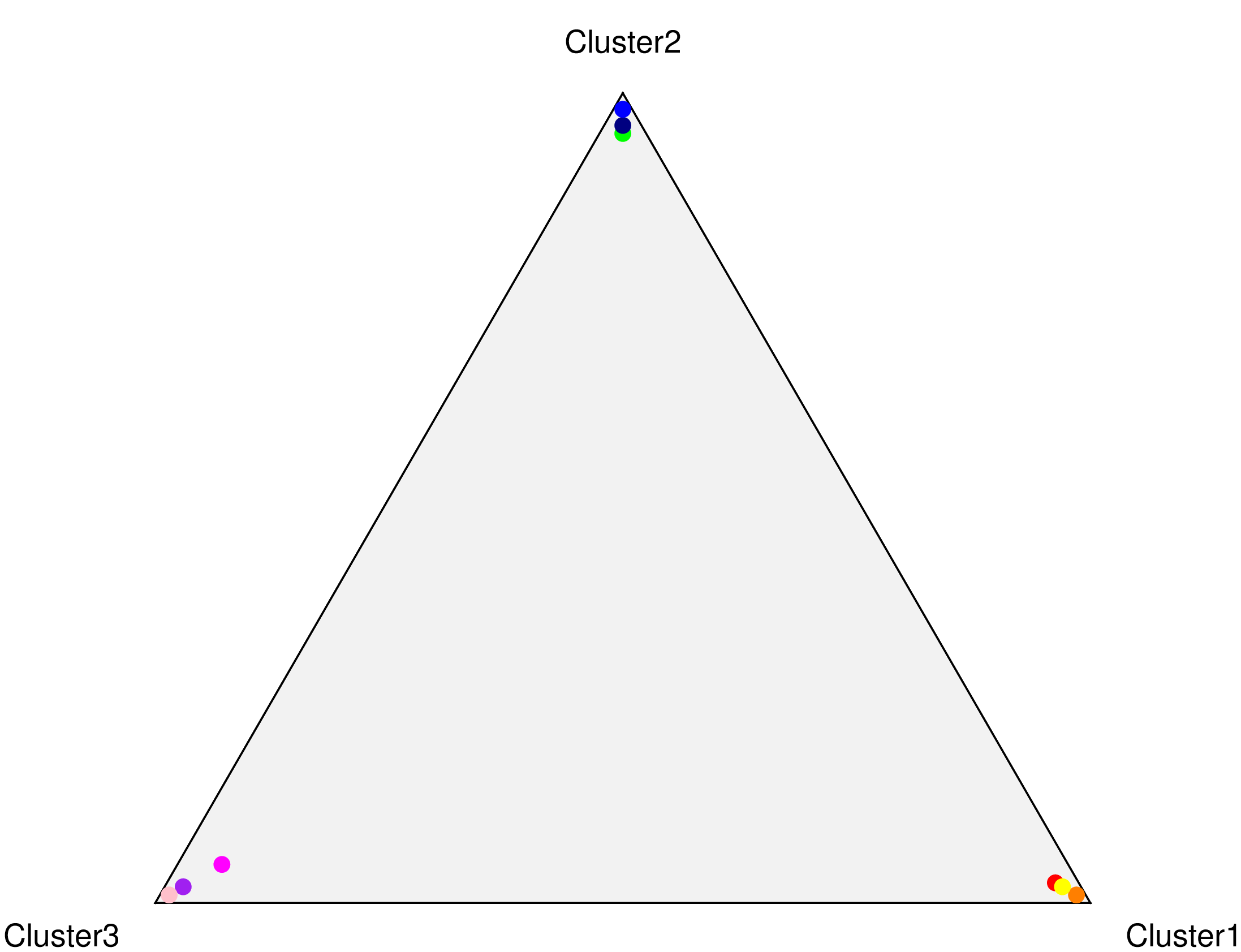}
        Even Low Fuzzy
    \end{minipage}
    \begin{minipage}[c]{0.32\linewidth}
        \centering
        \includegraphics[width=\linewidth]{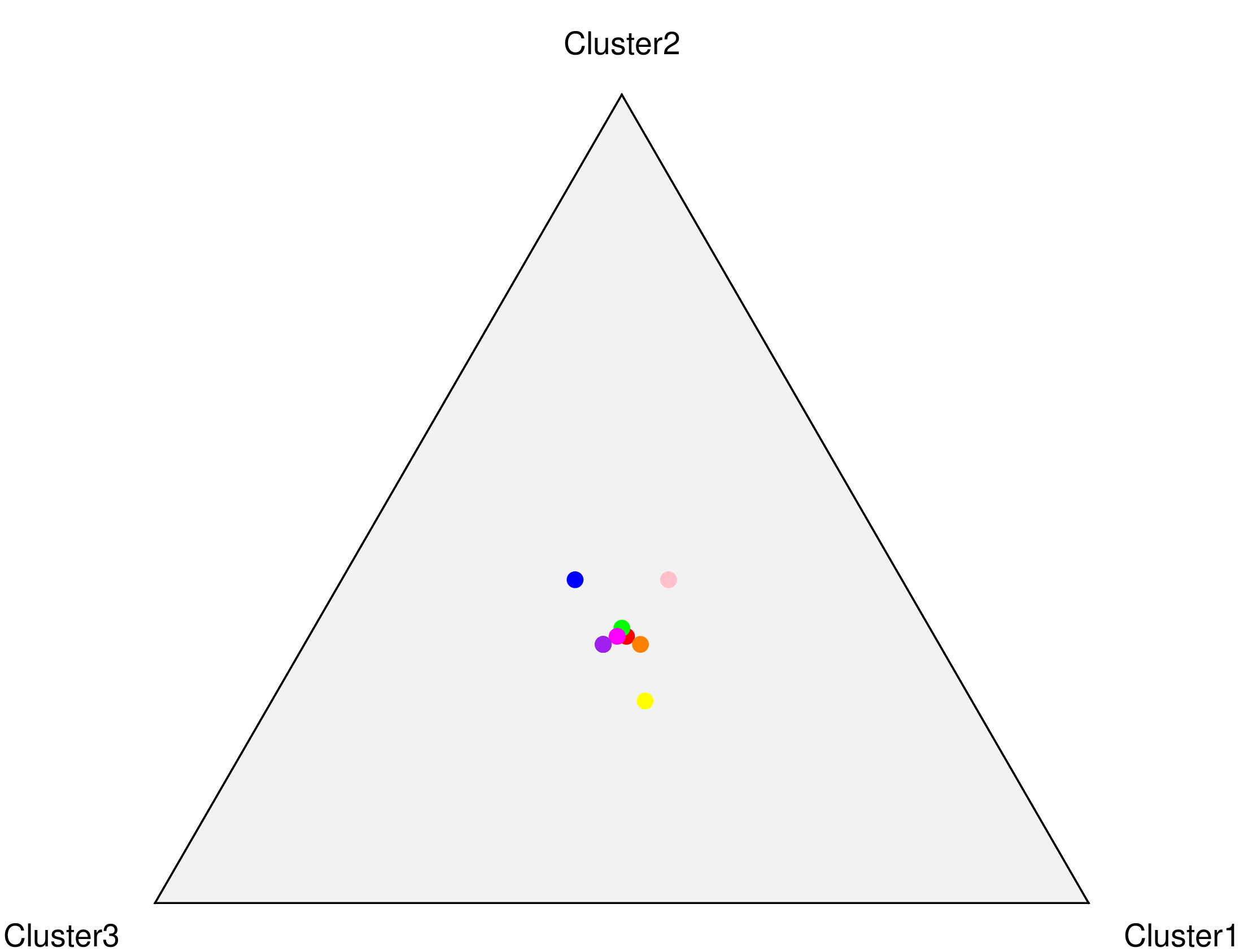}
        High Fuzzy
    \end{minipage}
    \caption{Toy Fuzzy clustering results}
    \label{fig:toyPoints}
\end{figure}

\begin{table}
\begin{minipage}[t]{0.45\linewidth}

    \begin{tabular}{|c|c|c|}
        \hline
        Id & $C_1$ & $C_2$ \\
        \hline
         1 &  Uneven LF & Even LF \\
         2 &  Uneven LF & HF \\
         3 &  Uneven LF & Uneven H \\
         4 &  Uneven LF & Even H \\
         5 &  Even LF & HF \\
         6 &  Even LF & Uneven H \\
         7 &  Even LF & Even H \\
         8 &  HF & Uneven H \\
         9 &  HF & Even H \\
         10 & Uneven H & Even H \\
         \hline
    \end{tabular}
    \caption{Comparison Ids. Low Fuzzy (LF), High Fuzzy (HF), Hard (H)}
\end{minipage}
\hfill
\begin{minipage}[t]{0.5\linewidth}
    \begin{tabular}{|c|c|c|c|c|}
    \hline
    NDC & Perm & Fit & Sym & Flat \\
    \hline
    3     & 3   &   3   &   3    &  3\\
    7   &   7   &   7   &   7   &   7\\
    2  &    1   &   10  &    8  &    2\\
    8  &    6   &   6   &   2    &  8\\
    1   &   10  &    4  &    6  &    1\\
    6   &   4   &   2   &   4   &   6\\
    10  &    2  &    8  &    9  &    10\\
    4   &   8  &    9   &   10  &    4\\
    5   &   5   &   5   &   5   &   5\\
    9   &   9  &    1   &   1   &   9\\
    \hline
    \end{tabular}
    \caption{ACI for each Comparison. Sorted Most similar to least similar.}
    \label{tab:toyExampleResults}
\end{minipage}
\end{table}

\begin{figure}
    \centering
    \includegraphics[width=\linewidth]{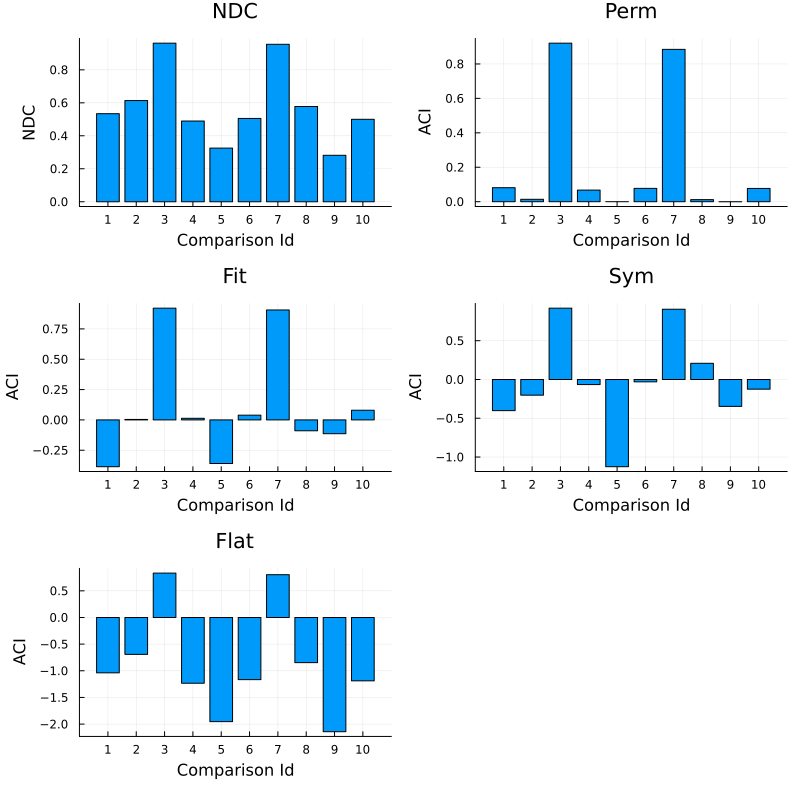}
    \caption{Toy Example ACI under each adjustment model.}
    \label{fig:toyExampleGraph}
\end{figure}

In this example, we make use of five synthetic cluster allocations of nine points into three clusters.
The three fuzzy clusterings are visualised in Figure \ref{fig:toyPoints}, with the other two being hard versions of UnevenLowFuzzy and EvenLowFuzzy.
Of the two even cluster allocations, each has 3 points assigned to each cluster, with the low fuzzy cluster allocation assigning a probability of 0.98 to one cluster, and 0.01 to each of the other two.
The two uneven cluster allocations have the same values, but 7 points are assigned to one cluster and one point to the two remaining.
Finally we have a high fuzzy clustering where every point has a membership vector close to $[0.33, 0.33, 0.33]$.
We calculate comparisons on all pairs of these clustering results using the implemented two-sided random models. 

As seen in Table \ref{tab:toyExampleResults}, comparisons 3 and 7 are consistently measured as the most similar.
These correspond to the Uneven Low Fuzzy \& Uneven Hard and the Even Low Fuzzy \& Even Hard comparisons respectively.
Given the data sets, it is clear that these should be considered as the most similar by far, and their consistently high similarity as seen in Figure \ref{fig:toyExampleGraph} serves as a evidence that all the models behave as expected.

Other than the obviously similar comparisons, the rest of the list is not stable between adjustment models.
For example, comparison 1 is reported as the third most similar under the permutation model, but the least similar under both the \fit and \sym models.

Using the \sym model, the adjustment tends to be dominated by concordance by disagreement.
This effect increases as the number of clusters grows, and does not reflect cluster allocations that have different sized clusters.
An example of where \sym may not be the best choice is in comparison 10.
The Uneven Low Fuzzy clustering has one group much larger than the others, and is being compared with a clustering that has even sized clusters.
It is highly unlikely that the model would produce such a pair of clustering results, and in general two algorithms that provide symmetrical sized groups are more similar than these two.
This behaviour is neither good nor bad, but must be considered when selecting an adjustment model.

Considering the interpretability of the results, Figure \ref{fig:toyExampleGraph} displays the ACI values for each of the models.
Of note, the permutation model has no negative values, indicating that all these comparisons are more similar than random.
However, the other models, especially \flatt suggest that many of the comparisons are worse than random.
In particular, comparison 1 has an ACI of 0.1 using the permutation model and between -0.3 and -1 for the three new models.
The clustering results from this comparison have very different cluster sizes and low fuzziness which we would not think to be very similar, and in fact are probably quite dissimilar.
But since the permutation model only considers clustering results with these exact attributes, all of the comparisons are going to be very dissimilar, and it is never considered that slightly less uneven clusters sizes could have occurred, which, once considered, significantly improve the measured similarity.

\subsection{Error Analysis}

Since the sampling method for calculating the expected index has a random element, it is necessary to analyze the variability of the computation.
We consider a range of synthetic clustering results, using a factorial scheme over number of clusters (2, 50), number of points (100, 1000), cluster size imbalance \footnote{This value indicate the proportion of clusters to which 80\% of points are assigned, evenly split among the clusters.} (0.8, 0.2), percentage of points that are randomized (0.5, 1.0), and precision of the Dirichlet used to generate random membership vectors (0.1, 1, 1.5).
In total there are 48 parameter settings, each of which was used to generate 10 pairs of cluster allocations.
The adjusted index was computed 100 times using a two sided comparison for each of the three Dirichlet models.

Since most studies will be looking for high, or at least positive, index values, we dropped all comparisons where all 100 of the computations were negative.
Removing these results does decrease the reported variability due to negative values generally having smaller denominators in Equation \ref{eq:ai}, but the results will be more accurate for common use cases including the other simulations in this section.
With the cleaned data set, we calculate the mean absolute error using the mean of the 100 values as the true expectation.
Since expectation is linear, the mean of the 100 values is equal to doing one computation with 1 000 000 000 samples, far more than is feasible for a reasonable run time (a few seconds vs. several minutes).

For all three models, over 99.5\% of computations have an absolute error of less than 0.01.
Furthermore, the maximum absolute error of the models is 0.02 for Fit and Sym and 0.002 for Flat.
These results give confidence that the computation method yields adjusted index values sufficiently close to the `true' value for accurate and meaningful comparison of cluster allocation similarity.

\subsection{A Factorial Benchmark}

In this section we analyze the behaviour of the proposed indices across a full factorial simulation.
Five cluster attributes were considered; number of clusters, number of points, cluster size imbalance, precision, and randomize rates with values in Table \ref{tab:paramSpace}.
Each run begins with two hard cluster allocations with perfect agreement.
The cluster imbalance setting was used to give 80\% of points to the specified proportion of clusters.
A Dirichlet parameterized using the precision variable, with 80\% of the precision weight given to the same cluster imbalance proportion of clusters.
The trial with 0 precision corresponds to using a categorical random variable instead of a Dirichlet to produce random hard membership vectors.
The randomize rate proportion of points from clustering allocations had their membership vectors replaced by i.i.d samples from the  Dirichlet (or Categorical) distribution.
For each set of parameters, 5 pairs of clusters were randomly generated.

In order to measure the trends associated with changing only one parameter, we consider the mean index value over all other parameters and over all runs.
The particular index values for this type of simulation can be difficult to interpret, since so many variations are considered in each case.
We will instead focus our analysis on the trends, which can be clearly seen and interpreted.

\begin{table}[]
    \centering
    \begin{tabular}{|l|l|}
        \hline
        Variable & Values\\
        \hline
        No. Clusters &  2, 4, 8, 16, 32, 64, 128\\
        No. Points   & 128, 256, 512, 1024, 2048, 4096, 8192\\
        Cluster Imbalance & 0.8, 0.6, 0.4, 0.2\\
        Precision & 0, 0.01, 0.1,  1, 1.5\\
        Randomize Rate & 0.2, 0.4, 0.6, 0.8, 1.0\\
        \hline
    \end{tabular}
    \caption{Parameter Space for the Factorial Simulation}
    \label{tab:paramSpace}
\end{table}

\begin{figure}[!h]
    \begin{minipage}[c]{0.32\linewidth}
    \centering
    \includegraphics[width=\linewidth]{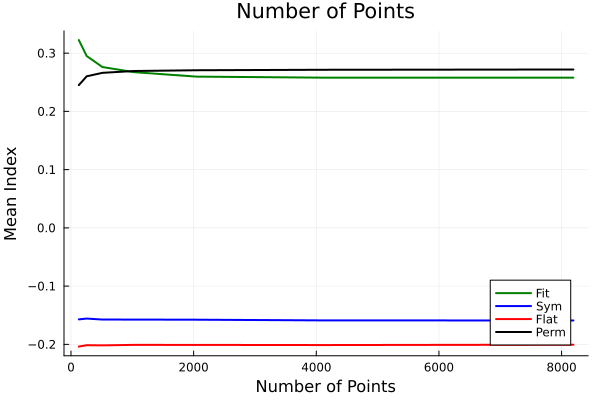}
    \end{minipage}
    \begin{minipage}[c]{0.32\linewidth}
        \centering
        \includegraphics[width=\linewidth]{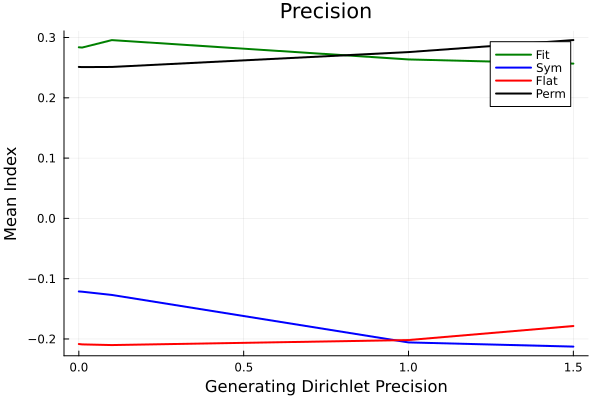}
    \end{minipage}
    \begin{minipage}[c]{0.32\linewidth}
        \centering
        \includegraphics[width=\linewidth]{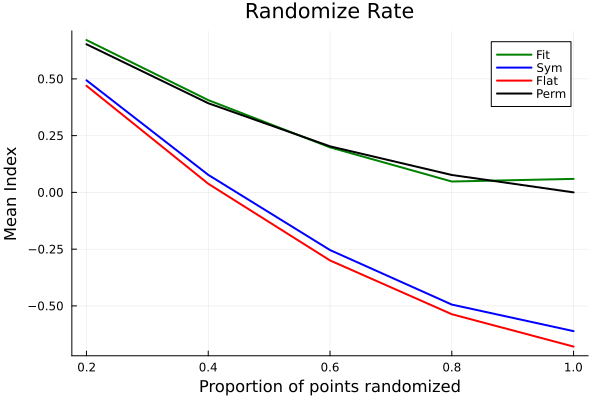}
    \end{minipage}
    \\
    \begin{minipage}[c]{0.45\linewidth}
        \includegraphics[width=\linewidth]{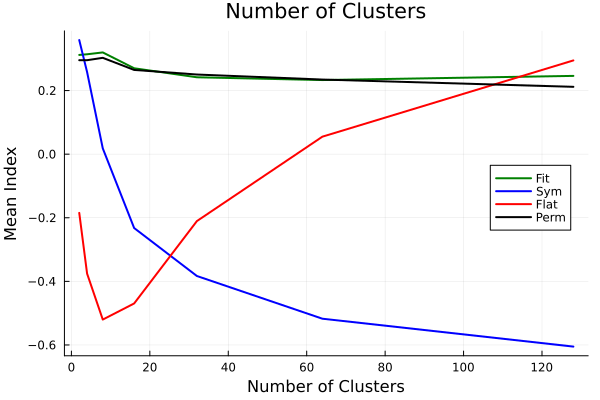}
    \end{minipage}
    \hfill
    \begin{minipage}[c]{0.45\linewidth}
        \includegraphics[width=\linewidth]{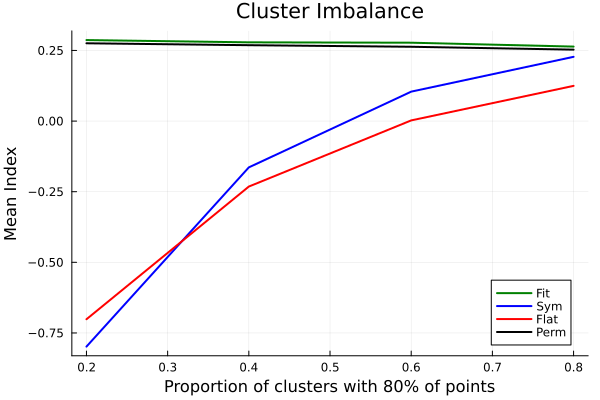}
    \end{minipage}
    \caption{Mean Index values of the two sided factorial benchmark. The mean of all runs and all parameters, except for the title parameter, were averaged at each x value.}
    \label{fig:TwoSidedGraphs}
\end{figure}

As expected, the \fit and \perm models behave quite similarly across all parameters, as seen in Figure \ref{fig:TwoSidedGraphs}, with a maximum difference of only 0.07.
However, we see different trends in the \sym and \flatt models.
On average, the \flatt model increases quickly with more clusters, while the \sym decreases quickly.
The \fit and \perm models both decrease quite slowly, an interesting result that conflicts with the purely hard case where the permutation model is known to increase with more clusters \citep{lei2017ground}.
Additionally, both the \flatt and \sym models increase on average as the cluster sizes become more balanced, while the \fit and \perm models remain unchanged.
Finally, as the only graph where we have an expected outcome, all four models decrease as the randomize rate increase, reflecting that less similar models do receive lower index values in all cases

\begin{figure}[!h]
    \begin{minipage}[c]{0.32\linewidth}
    \centering
    \includegraphics[width=\linewidth]{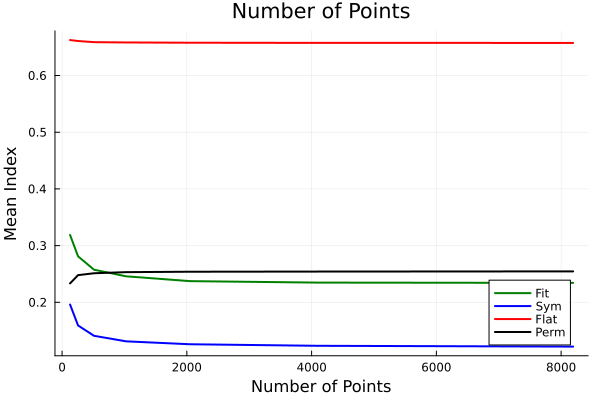}
    \end{minipage}
    \begin{minipage}[c]{0.32\linewidth}
        \centering
        \includegraphics[width=\linewidth]{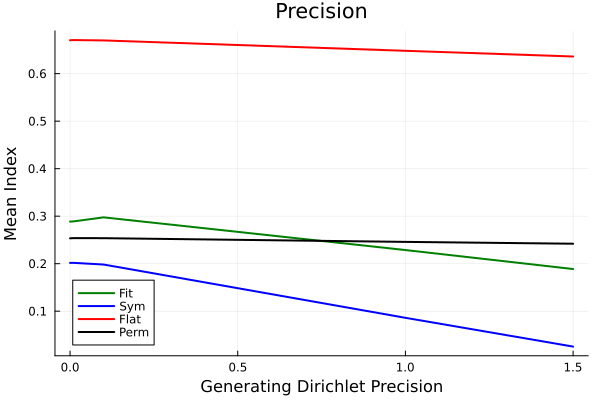}
    \end{minipage}
    \begin{minipage}[c]{0.32\linewidth}
        \centering
        \includegraphics[width=\linewidth]{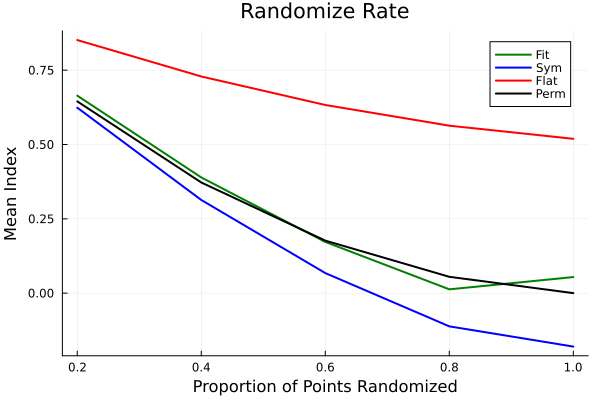}
    \end{minipage}
    \\
    \begin{minipage}[c]{0.45\linewidth}
        \includegraphics[width=\linewidth]{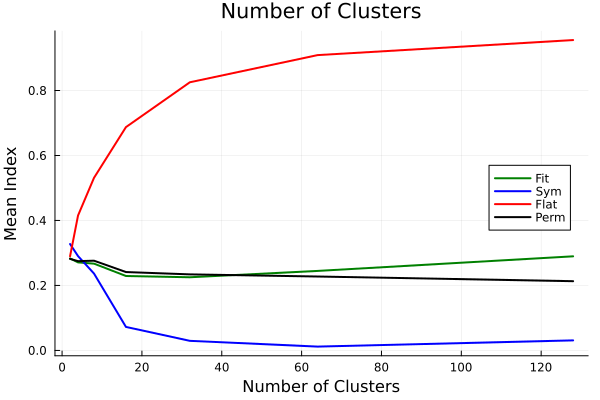}
    \end{minipage}
    \hfill
    \begin{minipage}[c]{0.45\linewidth}
        \includegraphics[width=\linewidth]{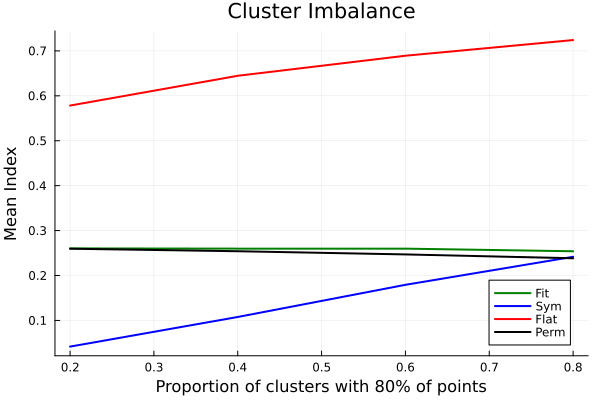}
    \end{minipage}
    \caption{Mean Index values of the one sided factorial benchmark. The mean of all runs and all parameters, except for the title parameter, were averaged at each x value.}
    \label{fig:OneSidedGraphs}
\end{figure}

We repeat the simulation, but only randomizing one clustering, and using the one-sided variants of the indexes.
This is the scenario most common for algorithm benchmarks; the reference clustering is fixed and many runs of an algorithm produce variants on a second clustering.
The trends for this scenario are much different than when both clusters are randomized, as seen in Figure \ref{fig:OneSidedGraphs}.
The trends for increasing number of clusters are quite complex.
The \fit, \sym, and \perm models all decrease from 2 to 16 clusters, but then \fit begins increasing while \sym and \perm remain constant.
\flatt sharply increases from 2 to 32 clusters, and then increases more slowly afterwards.
As before, \fit and \perm are constant with respect to cluster imbalance, while \sym and \flatt increase as clusters become more balanced.
Finally, as expected all indices are constant with respect to increasing number of points and decreasing with respect to more randomization.

\section{Conclusions}

In this paper, we defined \textit{Agreement-Concordance type Rand Extensions} for fuzzy clustering results and proposed three Dirichlet random models for index adjustment.
The Fit model is a generalization of the Categorical (Multinomial) model proposed by Morey and Agresti \citep{morey1984ARI}, with the Sym and Flat models used to bring the ideas of Gates and Ahn \citep{gates2017models} to the fuzzy realm.
Since the random model in index adjustment is used to set the baseline, a unified framework for randomness that has clear and explainable assumptions is of great benefit when communicating results of new or improved clustering algorithms.

The theory provided for the expected index of a random model are sufficient for any probability density function defined on a simplex, and various kernel density estimates or generalized Dirichlet distributions could be considered in future work.
These distributions could capture more information about covariance structure or levels of fuzziness.
Of course, if it is not fast to generate samples, numerical integration methods would also need to be explored.

Finally, we would like to reiterate the point from Gates and Ahn \citep{gates2017models} that one model is not better, or more accurate, than the others, and that model selection should be considered based on the assumptions used in a particular clustering algorithm.
Two-sided versions can be used when comparing pairs of clustering results.
The one-sided versions are useful when comparing competing models to a single reference set of allocations --- as is common for benchmarking purposes.

\bibliographystyle{unsrt}
\bibliography{references}

\begin{thebibliography}{10}

\bibitem{rand1971}
William~M Rand.
\newblock Objective criteria for the evaluation of clustering methods.
\newblock {\em Journal of the American Statistical association}, 66(336):846--850, 1971.

\bibitem{hubert1985ARI}
Lawrence Hubert and Phipps Arabie.
\newblock Comparing partitions.
\newblock {\em Journal of classification}, 2:193--218, 1985.

\bibitem{gates2017models}
Alexander~J Gates and Yong-Yeol Ahn.
\newblock The impact of random models on clustering similarity.
\newblock {\em Journal of Machine Learning Research}, 18:1--28, 2017.

\bibitem{morey1984ARI}
Leslie~C Morey and Alan Agresti.
\newblock The measurement of classification agreement: An adjustment to the rand statistic for chance agreement.
\newblock {\em Educational and Psychological Measurement}, 44(1):33--37, 1984.

\bibitem{mclachlan2019finite}
Geoffrey~J McLachlan, Sharon~X Lee, and Suren~I Rathnayake.
\newblock Finite mixture models.
\newblock {\em Annual review of statistics and its application}, 6:355--378, 2019.

\bibitem{mcnicholas2016mixture}
Paul~D McNicholas.
\newblock {\em Mixture model-based classification}.
\newblock CRC press, 2016.

\bibitem{andrews2022assessments}
Jeffrey~L Andrews, Ryan Browne, and Chelsey~D Hvingelby.
\newblock On assessments of agreement between fuzzy partitions.
\newblock {\em Journal of Classification}, 39(2):326--342, 2022.

\bibitem{brouwer2009}
Roelof~K Brouwer.
\newblock Extending the rand, adjusted rand and jaccard indices to fuzzy partitions.
\newblock {\em Journal of Intelligent Information Systems}, 32:213--235, 2009.

\bibitem{campagner2023frameworkFuzzyIndices}
Andrea Campagner, Davide Ciucci, and Thierry Den{\oe}ux.
\newblock A general framework for evaluating and comparing soft clusterings.
\newblock {\em Information Sciences}, 623:70--93, 2023.

\bibitem{hullermeier2011comparing}
Eyke Hullermeier, Maria Rifqi, Sascha Henzgen, and Robin Senge.
\newblock Comparing fuzzy partitions: A generalization of the rand index and related measures.
\newblock {\em IEEE Transactions on Fuzzy Systems}, 20(3):546--556, 2011.

\bibitem{d2021adjusted}
Antonio D’Ambrosio, Sonia Amodio, Carmela Iorio, Giuseppe Pandolfo, and Roberta Siciliano.
\newblock Adjusted concordance index: an extensionl of the adjusted rand index to fuzzy partitions.
\newblock {\em Journal of Classification}, 38:112--128, 2021.

\bibitem{steinley2004properties}
Douglas Steinley.
\newblock Properties of the hubert-arable adjusted rand index.
\newblock {\em Psychological methods}, 9(3):386, 2004.

\bibitem{steinley2018noteonrand}
Douglas Steinley and Michael~J Brusco.
\newblock A note on the expected value of the rand index.
\newblock {\em British Journal of Mathematical and Statistical Psychology}, 71(2):287--299, 2018.

\bibitem{steinley2016variance}
Douglas Steinley, Michael~J Brusco, and Lawrence Hubert.
\newblock The variance of the adjusted rand index.
\newblock {\em Psychological methods}, 21(2):261, 2016.

\bibitem{warrens2022understanding}
Matthijs~J Warrens and Hanneke van~der Hoef.
\newblock Understanding the adjusted rand index and other partition comparison indices based on counting object pairs.
\newblock {\em Journal of Classification}, 39(3):487--509, 2022.

\bibitem{jain2010data}
Anil~K Jain.
\newblock Data clustering: 50 years beyond k-means.
\newblock {\em Pattern recognition letters}, 31(8):651--666, 2010.

\bibitem{sundqvist2023mri}
Martina Sundqvist, Julien Chiquet, and Guillem Rigaill.
\newblock Adjusting the adjusted rand index: A multinomial story.
\newblock {\em Computational Statistics}, 38(1):327--347, 2023.

\bibitem{anderson2010comparing}
Derek~T Anderson, James~C Bezdek, Mihail Popescu, and James~M Keller.
\newblock Comparing fuzzy, probabilistic, and possibilistic partitions.
\newblock {\em IEEE Transactions on Fuzzy Systems}, 18(5):906--918, 2010.

\bibitem{minka2000mleDirichlet}
Thomas Minka.
\newblock Estimating a dirichlet distribution, 2000.

\bibitem{Julia-2017}
Jeff Bezanson, Alan Edelman, Stefan Karpinski, and Viral~B Shah.
\newblock Julia: A fresh approach to numerical computing.
\newblock {\em SIAM {R}eview}, 59(1):65--98, 2017.

\bibitem{Distributions.jl-2019}
Dahua Lin, John~Myles White, Simon Byrne, Douglas Bates, Andreas Noack, John Pearson, Alex Arslan, Kevin Squire, David Anthoff, Theodore Papamarkou, Mathieu Besançon, Jan Drugowitsch, Moritz Schauer, and other contributors.
\newblock {JuliaStats/Distributions.jl: a Julia package for probability distributions and associated functions}, July 2019.

\bibitem{lei2017ground}
Yang Lei, James~C Bezdek, Simone Romano, Nguyen~Xuan Vinh, Jeffrey Chan, and James Bailey.
\newblock Ground truth bias in external cluster validity indices.
\newblock {\em Pattern Recognition}, 65:58--70, 2017.

\end{thebibliography}

\appendix

\section{Proof of Stability} \label{app:A}

\stability*

\begin{proof}
    \begin{align*}
        &\left| \E_{\id{D_1}, \id{D_2}}[\conc] - \E_{D_1, D_2}[\conc] \right|\\
        =&\bigg| \int_{\simp_1}\int_{\simp_1} \int_{\simp_2} \int_{\simp_2} \id{f}(z_{11})\id{f}(z_{12})\id{g}(z_{21})\id{g}(z_{22}) \conc(z_{11}, z_{12}, z_{21}, z_{22}) dz_{11}dz_{12}dz_{21}dz_{22} \\
        &- \int_{\simp_1}\int_{\simp_1} \int_{\simp_2} \int_{\simp_2} f(z_{11})f(z_{12})g(z_{21})g(z_{22}) \conc(z_{11}, z_{12}, z_{21}, z_{22}) dz_{11}dz_{12}dz_{21}dz_{22} \bigg| \\
        \leq& \int_{\simp_1}\int_{\simp_1} \int_{\simp_2} \int_{\simp_2} \left| \id{f}(z_{11})\id{f}(z_{12})\id{g}(z_{21})\id{g}(z_{22}) - f(z_{11})f(z_{12})g(z_{21})g(z_{22}) \right| dz_{11}dz_{12}dz_{21}dz_{22} \\
        \leq & \int_{\simp_1} \left| \id{f}(z_{11}) - f(z_{11})\right| dz_{11} + \int_{\simp_1} \left| \id{f}(z_{12}) - f(z_{12})\right| dz_{12}\\
        &+ \int_{\simp_2} \left| \id{g}(z_{21}) - g(z_{21})\right| dz_{21} + \int_{\simp_2} \left| \id{g}(z_{22}) - g(z_{22})\right| dz_{22}\\
        \leq& \frac{\varepsilon}{4} + \frac{\varepsilon}{4} + \frac{\varepsilon}{4} + \frac{\varepsilon}{4}\\
        =& \varepsilon
    \end{align*}
\end{proof}

\end{document}